\def\year{2021}\relax
\documentclass[letterpaper]{article} 
\usepackage{aaai21}  
\usepackage{times}  
\usepackage{amsthm,amsmath,amssymb,bm}
\usepackage{helvet} 
\usepackage{courier}  
\usepackage[hyphens]{url}  
\usepackage{graphicx} 
\usepackage{graphicx}  
\usepackage{natbib}  
\usepackage{caption} 
\usepackage{subfigure}
\usepackage{lineno}
\usepackage{xr}
\newcommand{\nbb}{\mathbb{N}}

\newcommand{\bt}{\mathbf{t}}

\newcommand{\bw}{\mathbf{w}}
\newcommand{\fcal}{\mathcal{F}}
\newcommand{\ibb}{\mathbb{I}}
\newcommand{\xcal}{\mathcal{X}}
\newcommand{\wcal}{\mathcal{W}}

\newcommand{\hcal}{\mathcal{H}}

\newcommand{\zcal}{\mathcal{Z}}

\newcommand{\ycal}{\mathcal{Y}}

\newcommand{\bp}{\tilde{\phi}}

\newcommand{\ebb}{\mathbb{E}}

\newcommand{\bv}{\mathbf{v}}
\newcommand{\rbb}{\mathbb{R}}

\newcommand{\citealtt}[1]{\citeauthor{#1},\citeyear{#1}}
\newtheorem{theorem}{Theorem}
\newtheorem{lemma}[theorem]{Lemma}
\newtheorem{proposition}[theorem]{Proposition}

\theoremstyle{definition}
\newtheorem{definition}{Definition}

\newtheorem{assumption}{Assumption}
\newtheorem{example}{Example}
\theoremstyle{definition}

\newtheorem{remark}{Remark}
\title{Fine-grained Generalization Analysis of Vector-valued Learning\footnote{
    To appear in AAAI 2021.}
\author{
    Liang Wu \textsuperscript{\rm 1}, Antoine Ledent \textsuperscript{\rm 2}, Yunwen Lei \textsuperscript{\rm 3, 2} and Marius Kloft \textsuperscript{\rm 2}}
}

\affiliations{

    \textsuperscript{\rm 1}Center of Statistical Research, School of Statistics, Southwestern University of Finance and Economics, Chengdu 611130, China\\
    \textsuperscript{\rm 2}Department of Computer Science, TU Kaiserslautern, 67653 Kaiserslautern, Germany\\
    \textsuperscript{\rm 3}School of Computer Science, University of Birmingham, Birmingham B15 2TT, United Kingdom
}

\begin{document}

\maketitle

\begin{abstract}
  Many fundamental machine learning tasks can be formulated as a problem of learning with vector-valued functions, where we learn multiple scalar-valued functions together. Although there is some generalization analysis on different specific algorithms under the empirical risk minimization principle, a unifying analysis of vector-valued learning under a regularization framework is still lacking. In this paper, we initiate the generalization analysis of regularized vector-valued learning algorithms by presenting bounds with a mild dependency on the output dimension and a fast rate on the sample size. Our discussions relax the existing assumptions on the restrictive constraint of hypothesis spaces, smoothness of loss functions and low-noise condition. To understand the interaction between optimization and learning, we further use our results to derive the first generalization bounds for stochastic gradient descent with vector-valued functions. We apply our general results to multi-class classification and multi-label classification, which yield the first bounds with a logarithmic dependency on the output dimension for extreme multi-label classification with the Frobenius regularization. As a byproduct, we derive a Rademacher complexity bound for loss function classes defined in terms of a general strongly convex function.
\end{abstract}

\section{Introduction}

In machine learning, we often encounter learning tasks involving vector-valued prediction functions \citep{alvarez2012kernels,xu2019survey}. As examples consider the following. In multi-class classification (MCC), we aim to assign each instance to a single label class \citep{mohri2012foundations,crammer2002algorithmic}. In multi-label classification (MLC), each instance may be annotated with one or multiple class labels \citep{zhou2012multi,yu2014large}. In both cases, we build one scalar-valued prediction function per class. Together these functions form a vector-valued predictor with the output dimension being the total number of label classes. These two important learning tasks have found wide applications in real systems, where they are used, for instance, for image and video annotation, face classification, and query/keyword suggestion~\citep{yu2014large}.
Another popular learning setting where we encounter vector-valued prediction functions is multi-task learning (MTL). Here we build, for each task, a distinct predictor. The benefit of learning these predictors together is that of exploiting a shared hidden structure in these tasks \citep{zhang2017survey,maurer2016bounds,yousefi2018local,argyriou2007spectral,ciliberto2017consistent}. The more these tasks are related, the larger the structure and the benefit of learning them together. We refer to problems of learning a vector-valued prediction function---such as the above ones---as \emph{vector-valued learning problems} \citep{micchelli2005learning,8263148,xu2019survey,alvarez2012kernels,lei2015theory}.

An important measure of the quality of vector-valued learning models is their generalization performance, i.e., their ability to generalize their empirical behavior on training examples to unseen test data. As a central topic in statistical learning theory, generalization analysis has received a lot of attention. Other than providing an intuitive understanding of how different parameters  affect the learning performance, generalization analysis is also effective in designing novel learning machines \citep{cortes2013learning}.

Unlike traditional binary classification problems, a distinguished property of vector-valued learning is that the output dimension plays an important role in the analysis~\citep{reddi2019stochastic}. This is especially the case for problems with a huge output dimension, which are becoming more and more ubiquitous in the big data era. One such example is eXtreme Classification \citep[XC;][]{bengio2019extreme}. Here we deal with multi-class and multi-label problems involving an extremely large total number of potential class labels~\citep{jain2019slice}. Generalization bounds with an emphasis on the output dimension were developed for specific vector-valued learning problems such as MCC \citep{zhang2004statistical,lei2015multi,guermeur2017lp,li2018multi,musayeva2019rademacher} or MLC~\citep{yu2014large,liu2018dual,shen2018compact,xu2016local,khandagale2020bonsai}.

Recently, \citet{maurer2016bounds,li2019learning} initiated the study of the general framework of vector-valued learning. These studies exhibit the following limitations.
First, they exploit the Lipschitz continuity of loss functions with respect to (w.r.t.) the Euclidean norm, while typical loss functions occurring in vector-valued learning problems are Lipschitz continuous w.r.t. the $\ell_\infty$-norm \citep{lei2019data}, with a comparable Lipschitz constant as for the Euclidean norm (note $\ell_\infty$-norm can be significantly smaller than the Euclidean norm). This mismatch between generalization analysis and Lipschitz continuity induces, for standard Euclidean regularization, a square-root dependency on the number of components (that is, the number of classes or tasks in MCC/MLC and MTL, respectively) \citep{maurer2016bounds,li2019learning}.
Second, there is the following conceptual mismatch between algorithms and theory. While the theory is developed for empirical risk minimization (ERM) in a constrained space~\citep{maurer2016bounds,li2019learning,reeve2020optimistic}, in practice, regularization schemes are used, which are oftentimes easier to solve \citep{yu2014large,lei2015multi,li2018multi,li2019learning}.
Third, the existing generalization analysis fails to take into account the computational properties of the algorithm, which is important to understand the interaction between optimization and learning.
Lastly, most existing studies are limited to the classic regime of ``slow'' rates ($\Omega(n^{-\frac{1}{2}})$, where $n$ is the sample size).
In order to achieve so-called \emph{fast rates} \citep{bartlett2005local}, they require restrictive assumptions, such as the smoothness of loss function \citep{reeve2020optimistic}, the capacity assumption on the hypothesis space, or the existence of a model with vanishing errors. \\[2pt]
\indent In this paper, we address all the above issues. Our contributions are as follows. \\[2pt]
\indent 1. We show the first generalization bounds for general vector-valued learning using a \emph{regularization} framework, thus removing the gap between the algorithm that is analyzed in theory and the one that is considered in practice. \\[2pt]
\indent 2. Not only do we analyze a more realistic model, we dramatically improve the best known dependency of bounds for general vector-valued learning (in either framework). For instance for standard Frobenius regularization, we drop the dependency on the number of components in the model from $\sqrt{c}$ to $\log{c}$.
\\[2pt]
\indent 3. In MLC, the components are the classes. Thus our result establishes guarantees that scale logarithmic in the number of classes. This is remarkable because the previously best result for MLC scaled square root in the number of classes. Thus, for the first time, we establish non-void theory for extreme multi-label classification (yet, as mentioned in Point 1, our analyzed algorithm is more realistic). Note that MLC is the by far most common scenario in XC. \\[2pt]
\indent 4. Our results apply also to the fast-rate regime, that is, the learning scenario where bounds enjoy a fast decay on $n$. In this regime, our analysis improves not only the best known rates in $c$ (see Point 2) over previous work, but lifts also assumptions employed therein on the smoothness of the loss function and hypothesis space.


\section{Related Work\label{sec:work}}
Here we survey the related work on vector-valued learning.

There is a large body of work on the generalization analysis of MCC~\citep{lei2019data,thrampoulidis2020theoretical}, based on various capacity measures: e.g., covering numbers~\citep{zhang2004statistical,lei2019data}, the fat-shattering dimension~\citep{guermeur2017lp}, and (local) Rademacher complexities~\citep{mohri2012foundations,lei2015multi,maurer2016vector,cortes2016structured,li2018multi,maximov2018rademacher,musayeva2019rademacher,musayeva2018sharper,deshmukh2019generalization}. Unlike binary classification problems, the number of classes plays here an important role in the generalization performance. Until recently, the coupling among the class components, while exploited by most practical multi-class algorithms \citep{crammer2002algorithmic}, was ignored by generalization bounds. As a result, they exhibited at least a linear dependency on the number of classes~\citep{mohri2012foundations}. Subsequently, several works aimed to improve this dependency through structural results on Gaussian and Rademacher complexities that exploit the coupling among classes, first achieving a square root  \citep{guermeur2017lp,maurer2016vector,lei2015multi} and later on a logarithmic dependency  \citep{lei2019data}. For the low-noise regime, \citet{li2018multi} show a fast-rate generalization bound based on local Rademacher complexities.

There is far less work on the theoretical analysis of MLC. The consistency of MLC with different loss functions was studied in \citet{gao2011consistency}. For decomposable loss functions involving the specific least-squares loss, generalization bounds based on the Rademacher complexity were derived in \citet{yu2014large}. The best known dependency on the output dimension is square root and was shown in \citet{liu2018dual,wu2020multi}.
\citet{xu2016local} bounded the local Rademacher complexity of MLC, which motivated the authors to study a novel MLC algorithm based on the tail-sum of singular values of the predictors. 

Generalization analysis for general vector-valued learning algorithms was initiated by \citet{maurer2016bounds} and put forward by \citet{li2019learning}. However, their analysis implies generalization bounds with a square-root dependency on the output dimension for reasonable regularizers such as those based on the Frobenius norm.
This is because they work with the Lipschitz continuity w.r.t. the Euclidean norm. In the present work, we consider the infinity norm instead.
Notice that the existing results for MCC, MLC or general vector-valued learning algorithms are mainly established for ERM in a constrained hypothesis space~\citep{maurer2016bounds,li2019learning,reeve2020optimistic}. As a comparison, there is scarce work on vector-valued learning based on regularization, while in practice this is the mostly used scheme~\citep{crammer2002algorithmic,lei2015multi,yu2014large}.

\section{Problem Formulation and Results\label{sec:main}}
\subsection{Problem formulation}
We describe here the framework of vector-valued learning with regularization.
Let $\rho$ be a probability measure defined over a sample space $\zcal=\xcal\times\ycal$, where $\xcal\subseteq\rbb^d$ is an input space and $\ycal$ is an output space ($d$ is the input dimension). Let $S=\{z_1,\ldots,z_n\}\in\zcal^n$ be the training dataset drawn independently from $\rho$.
For vector-valued learning, we aim to build a vector-valued predictor $h:\xcal\mapsto\rbb^c$, i.e., $h=(h_1,\ldots,h_c)$ with $h_j:\xcal\mapsto\rbb$. We denote by $c$ the output dimension. We consider non-parametric learning in a reproducing kernel Hilbert space $\hcal_K$ associated with a Mercer kernel $K:\xcal\times\xcal\mapsto\rbb$. Let $\phi:\xcal\mapsto\hcal_K$ be the corresponding feature map, i.e., $K(x,x')=\langle\phi(x),\phi(x')\rangle$ for all $x,x'\in\xcal$ with $\langle\cdot,\cdot\rangle$ being the inner product. Consider the hypothesis space \[\wcal=\{\bw=(\bw_1,\ldots,\bw_c)\in\hcal_K^c\}.\]
We consider vector-valued predictors of the form \[h^{\bw}(x)=\big(h^{\bw}_1(x),\ldots,h^{\bw}_c(x)\big),\] where $\bw\in\wcal$ and $h^{\bw}_j(x)=\langle\bw_j,\phi(x)\rangle$.
Note that if $\phi(x)=x$, then the model $h^{\bw}$ can be characterized by a matrix $\bw$ in $\rbb^{d\times c}$ with $\bw_j$ being the $j$-th column.
The performance of $h^{\bw}$ on a single example $z$ is measured by a loss function $\ell:\wcal\times\zcal\mapsto\rbb_+$.
An effective approach to building a model is to learn with regularization, where we build an objective function $F_S:\wcal\mapsto\rbb_+$ by
\begin{equation}\label{F-S}
F_{S}(\bw)=\frac{1}{n}\sum_{i=1}^{n}\ell(\bw;z_i)+r(\bw).
\end{equation}
The first term characterizes the empirical behavior of a model $\bw$ on the training examples $S$, while the second term $r:\wcal\mapsto\rbb_+$ is a regularizer.
The predictor is then established by minimizing the objective function over the hypothesis space (regularized risk minimization, RRM), i.e., $\bw_{S}=\arg\min_{\bw\in\wcal}F_{S}(\bw)$. The generalization behavior of a model on a testing example can be measured by the regularized risk defined by
$F(\bw)=\ebb_Z\big[\ell(\bw;Z)\big]+r(\bw)$, where $\ebb_Z$ denotes the expectation w.r.t. $Z$. As we will show in applications, this framework of vector-valued learning covers important learning tasks such as
MCC and MLC by considering specific instantiations of the output space, loss functions and regularizers. For any $k\in\nbb$, let $[k]=\{1,\ldots,k\}$.

To study high-probability bounds for excess regularized risks of $\bw_S$,  we introduce some necessary assumptions.
The first assumption is the Lipschitz continuity of loss functions.
\begin{assumption}[Lipschitz continuity\label{ass:lip}]
We assume $\ell$ satisfies a Lipschitz continuity w.r.t. the infinity norm as follows
\begin{equation}\label{lipschitz}
  \big|\ell(\bw;z)-\ell(\bw';z)\big|\leq L\|h^{\bw}(x)-h^{\bw'}(x)\|_\infty,
\end{equation}
where $L>0$ and $\|\bt\|_\infty=\max_{j\in[c]}|t_j|$ for $\bt=(t_1,\ldots,t_c)$.
\end{assumption}
A notable property is that we consider the Lipschitz continuity w.r.t. the infinity-norm instead of the Euclidean norm in the literature~\citep{li2018multi,li2019learning}. Although these norms are equivalent, the involved Lipschitz constant can differ up to a factor of $\sqrt{c}$ which plays an important role in the generalization behavior if the output dimension is large. Therefore, the L-Lipschitz condition w.r.t. infinity norm is much stronger than that w.r.t. the Euclidean norm. Fortunately, popular loss functions in MLC and MCC actually satisfy the more restrictive condition of the Lipschitz continuity w.r.t. infinity-norm, where the Lipschitz constant is independent of $c$. This is why we can exploit the restrctive assumption on infinity-norm to develop a bound with a better dependency on $c$, which is an advantage over the analysis w.r.t. Euclidean norm~\citep{li2019learning}.

Our second assumption is the (strong) convexity of loss functions and regularizers~\citep{sridharan2009fast,kakade2012regularization}.
\begin{assumption}\label{ass:r}
  We assume $\ell$ is convex w.r.t. the first argument. We also assume $r$ is $\sigma$-strongly convex w.r.t. a norm $\|\cdot\|$, i.e., for all $\bw,\bw'\in\wcal$
  \[
  r(\bw)\geq r(\bw')+\langle\bw-\bw',r'(\bw')\rangle+\frac{\sigma}{2}\|\bw-\bw'\|^2,
  \]
  where $r'(\bw')$ denotes a subgradient of $r$ at $\bw'$.
\end{assumption}
A popular strongly convex regularizer is $r(\bw)=\frac{\sigma}{2}\|\bw\|_{2,p}^2$, where $\|\bw\|_{2,p}=\big(\sum_{j=1}^{c}\|\bw_j\|_2^p\big)^{\frac{2}{p}}$ is the $\ell_{2,p}$ norm ($p\geq1$).
Here $\|\cdot\|_2$ denotes the norm in $\hcal_K$ induced by $\langle\cdot,\cdot\rangle$.
It is known that this regularizer is $\sigma(p-1)$-strongly convex w.r.t. $\|\cdot\|_{2,p}$ for $p\in(1,2]$~\citep{kakade2012regularization}. Another popular regularizer is $r(\bw)=\frac{\sigma}{2}\|\bw\|_{S_p}^2$ for $\bw\in\rbb^{d\times c}$ where $\|\bw\|_{S_p}$ is the Shatten-$p$ norm~\citep{kakade2012regularization}. This regularizer is $(p-1)\sigma$-strongly convex w.r.t. $\|\cdot\|_{S_p}$ for $p\in(1,2]$.
Furthermore, we always assume $\|\bw\|\geq\|\bw\|_{2,\infty}$ for all $\bw\in\wcal$. This is a very mild assumption and is satisfied for all the norm considered in this paper. Finally, we assume $\sup_{x\in\xcal}\|\phi(x)\|_2\leq\kappa$.


\subsection{Main Results}
We now present our results.
We use Rademacher complexity to measure the capacity of hypothesis spaces.
\begin{definition}[\label{def:rademacher}Rademacher complexity]
  Let $\hcal$ be a class of real-valued functions defined over a space ${\zcal}$ and $S=\{{z}_i\}_{i=1}^n\in{\zcal}^n$.
  The \emph{empirical Rademacher complexities} of $\hcal$ with respect to $S$ is defined as
  \[\mathfrak{R}_{S}(\hcal)=\ebb_{\bm{\epsilon}}\big[\sup_{h\in \hcal}\frac{1}{n}\sum_{i=1}^n\epsilon_ih({z}_i)\big],\]
  where $\epsilon_1,\ldots,\epsilon_n$ are independent Rademacher variables, i.e., $\epsilon_i$ takes an equal probability of being either $1$ or $-1$.
\end{definition}
Our first result is an upper bound on the Rademacher complexity of loss function classes. This result is general in the sense that we define hypothesis spaces in terms of a general strongly convex function $\tau(\bw)=F(\bw)-F(\bw^*)$, where $\bw^*=\arg\min_{\bw\in\wcal}F(\bw).$
For any $\Lambda>0$, we denote
\begin{equation}\label{wcal}
  \wcal_\Lambda=\{\bw\in \wcal:F(\bw)-F(\bw^*)\leq \Lambda\}
\end{equation}
and the associated class of loss functions
\begin{equation}\label{fcal}
\fcal_\Lambda=\big\{\bw\mapsto\ell(\bw;z):\bw\in\wcal_\Lambda\big\}.
\end{equation}

\begin{theorem}[Rademacher complexity bound\label{thm:rademacher}]
  Let Assumptions \ref{ass:lip} and \ref{ass:r} hold. Then there exists a constant $C_1$ independent of $n,c,L$ and $\Lambda$ such that
  \[
  \mathfrak{R}_S(\fcal_\Lambda)\leq \frac{C_1L\sqrt{2\Lambda}\widetilde{B}\log^{2}(nc)}{\sqrt{n\sigma}},
  \]
  where $\widetilde{B}=\sup_{(x,j)}\|\bp_j(x)\|_*$ and $\|\cdot\|_*$ is the dual norm of $\|\cdot\|$ in $\wcal$.
  Here for any $x\in\xcal$ and $j\in[c]$ we use the notation
  $\bp_j(x):=\big(\underbrace{0,\ldots,0}_{j-1},\phi(x),\underbrace{0,\ldots,0}_{c-j}\big)\in \hcal_K^c.$
\end{theorem}
With different loss functions and regularizers, Theorem \ref{thm:rademacher} immediately implies specific Rademacher complexity bounds for different vector-valued learning problems. It is worth mentioning that the upper bound enjoys a logarithmic dependency on the output dimension (we treat $\Lambda$ as a constant which is problem-dependent and should be tuned by cross validation in practice). Therefore, this result is particularly useful for large-scale learning problems. This mild dependency is derived by exploiting the Lipschitz continuity of loss functions w.r.t. $\|\cdot\|_\infty$ and a structural result to capture this Lipschitz continuity. The proof of Theorem \ref{thm:rademacher} is given in Section \ref{sec:proof-rademacher} in the Appendix. 
\begin{remark}
  We now compare this result with the existing Rademacher complexity bounds for loss function classes in vector-valued learning problems. Initially, the Rademacher complexity bounds (Lemma 8.1 in \citet{mohri2012foundations}) failed to capture the coupling of different components reflected by the constraint (e.g., $\|\bw\|\leq\Lambda$ for some $\|\cdot\|$) and therefore presented a crude dependency on the output dimension. These results are improved in \citet{lei2015multi} and \citet{maurer2016vector} by exploiting the Lipschitz continuity of loss functions w.r.t. $\ell_2$ norm. Very recently, the Lipschitz continuity w.r.t. $\ell_\infty$ norm is also considered in the literature~\citep{lei2019data,foster2019ell}. However, the analysis in \citet{foster2019ell} failed to exploit the coupling among components while the analysis in \citet{lei2019data} only exploited this coupling enforced by some specific norms (case by case investigation is required).
   For example, they require Khintchine inequalities for vectors and matrices to handle $(2,p)$-norm and Schatten $p$-norm, respectively. As a comparison, Theorem \ref{thm:rademacher} provides a more general result where $\bw$ is constrained via a strongly convex function. A nice property is that Theorem \ref{thm:rademacher} treats $(2,p)$-norm and Schatten $p$-norm exactly the same, and does not need case-by-case discussions. Moreover, Theorem \ref{thm:rademacher} also applies to other regularization schemes, e.g., learning with entropic regularizer in a probability simplex.
\end{remark}

We now present upper bounds on the excess regularized risk for vector-valued learning.
We use a variant of the big-O notation $\widetilde{O}$ to hide any logarithmic factor. The proof is given in Section \ref{sec:proof-main} in the Appendix.

\begin{theorem}[\label{thm:main}Regularized risk bound for RRM]
Let Assumptions \ref{ass:lip} and \ref{ass:r} hold. Let $\delta\in(0,1)$.
With probability at least $1-\delta$ Eq. \eqref{main-a} holds uniformly for all $\bw\in\wcal$
\begin{multline}\label{main-a}
  F(\bw)-F(\bw^*) =  \max\big\{F_S(\bw)-F_S(\bw^*),1/(n\sigma)\big\}+\\ \widetilde{O}\Big(\Big(\frac{\big(F(\bw)-F(\bw^*)\big)\big(\log(1/\delta)+\widetilde{B}\big)}{n\sigma}\Big)^{\frac{1}{2}}\Big).
\end{multline}
In particular, the following inequality holds with probability at least $1-\delta$
\begin{equation}\label{main-b}
F(\bw_S)-F(\bw^*)=\widetilde{O}\Big(\frac{\log(1/\delta)+\widetilde{B}}{n\sigma}\Big).
\end{equation}
\end{theorem}

Eq. \eqref{main-a} applies uniformly to all $\bw\in\wcal$ while \eqref{main-b} applies to the RRM scheme. Both of these upper bounds admit a very mild dependency on the output dimension. Unlike ERM which generally implies slow rates $O(1/\sqrt{n})$~\citep{lei2019data}, we establish fast rates of the order $\widetilde{O}(1/n)$ for RRM by leveraging the strong convexity of the objective function. A notable property of this result is its generality. It requires only Lipschitz continuity of loss functions and strong convexity of objective functions to develop meaningful generalization bounds for any vector-valued learning methods. It does not impose a capacity assumption on hypothesis spaces or a low-noise condition for fast generalization bounds as was required in the existing treatments~\citep{li2018multi,li2019learning}.

\begin{remark}
  Generalization error bounds with an implicit dependency on the output dimension were recently derived for ERM with vector-valued functions~\citep{li2019learning}. The discussions there imposes a constraint in terms of either $\|\bw\|_{2,1}$ or $\|\bw\|_{S_1}$ on the hypothesis space.
  As a comparison, we get a logarithmic dependency by considering a regularizer involving a much milder norm $\|\cdot\|_{2,2,}$ (note $\|\cdot\|_{2,1}$ can be as large as $\sqrt{c}\|\cdot\|_{2,2}$). Indeed, the analysis in \citet{li2019learning} would imply a generalization bound with a square-root dependency on $c$ if the constraint $\|\bw\|_{2,1}\leq\Lambda$ there is replaced by $\|\bw\|_{2,2}\leq \Lambda$. We achieve this improvement by exploiting the Lipschitz continuity of loss functions w.r.t. the infinity-norm instead of w.r.t. the Euclidean norm. Although the Lipschitz contintuity w.r.t. infinity-norm is a much stronger assumption than that w.r.t. Euclidean norm, it is satisfied by popular loss functions with the Lipschitz constant independent of $c$ (cf. Section on Applications). Our analysis is able to fully exploit this stronger assumption and therefore gets a tighter dependency on $c$. The use of strong convexity allows for a tighter dependency on $n$.
\end{remark}

\begin{table*}[htbp]
\caption{Generalization bounds for MCC in a Frobenius learning framework (either constraint $\|W\|_F\leq 1$ or regularizer $\|W\|_F^2$)\label{tab:mcc}}
\begin{tabular}{|c|c|c|c|c|}
  \hline
  Bound & Lipschitz Continuity & Additional Assumption & Method & Reference \\ \hline
  $c/\sqrt{n}$ & $\ell_1$-norm & No & ERM & \citet{kuznetsov2014multi} \\ \hline
  $\sqrt{c/n}$ & $\ell_2$-norm & No & ERM & \citet{lei2015multi} \\ \hline
  $\sqrt{c}\log^3(n)/n$ & $\ell_2$-norm & smoothness and low noise & ERM & \citet{li2018multi} \\ \hline
  $\log^2(nc)/\sqrt{n}$ & $\ell_\infty$-norm & No & ERM & \citet{lei2019data} \\ \hline
  $\sqrt{c}/n$ & $\ell_2$-norm  &   decay of singular values & ERM & \citet{li2019learning} \\ \hline
  $\log^3(nc)/n$ & $\ell_\infty$-norm & strong convexity & RRM/SGD &  This work \\
  \hline
\end{tabular}
\end{table*}

Theorem \ref{thm:main} establishes excess risk bounds for $\bw_S$ under RRM, which is non-constructive in the sense that it does not tell us how the model $\bw_S$ is built from the data. In the following theorem, we will consider a constructive algorithm called SGD, which is a very popular optimization algorithm especially useful for large data analysis due to its simplicity and efficiency. To study its excess risk bounds, we need to consider both statistical and computational properties as well as the trade-off realized by early-stopping. We give the proof in Section \ref{sec:proof-sgd} (Appendix).
\begin{theorem}[Regularized risk bound for SGD\label{thm:sgd}]
  Assume $\ell(\bw;z)$ takes the form $\ell(\bw;z)=\psi(h^{\bw}(x),y)$ with $\psi:\rbb^c\times\ycal\mapsto\rbb$ being a convex function w.r.t. the first argument. Let Assumption \ref{ass:lip} hold and $r(\bw)=\frac{\sigma}{2}\|\bw\|_{2,2}^2$. Let $f(\bw;z)=\ell(\bw;z)+r(\bw)$. Let $\bw_1=0\in\wcal$ and $\{\bw_t\}_{t\in\nbb}$ be the sequence produced by SGD, i.e.,
  \begin{equation}\label{SGD}
    \bw_{t+1}=\bw_t-\eta_tf'(\bw_t;z_{i_t}),
  \end{equation}
  where $\eta_t=1/(t\sigma)$ and $\{i_t\}_t$ is independently drawn from the uniform distribution over $[n]$. Then for any $\delta\in(0,1)$ with probability at least $1-\delta$ the following holds:
  \begin{equation}\label{thm-sgd-res}
  F(\bw_T)-F(\bw^*) =\widetilde{O}\Big(\frac{\log(1/\delta)}{\min\{n,T\}\sigma}\Big).
  \end{equation}
  If $T$ is of the order of $n$ then with probability at least $1-\delta$,
  \begin{equation}\label{thm-sgd-res-b}
    F(\bw_T)-F(\bw^*)=\widetilde{O}\Big(\frac{\log(1/\delta)}{n\sigma}\Big).
  \end{equation}
\end{theorem}

\begin{remark}
  According to Theorem \ref{thm:sgd}, there is no need to run more iterations once $T$ is of the order of $n$ since further computation would not further improve the generalization performance. In other words, a computationally efficient method is to early-stop SGD after a constant number of passes over the data.
  To our best knowledge, this is the first result on the generalization analysis of SGD for vector-valued learning algorithms.
\end{remark}

\section{Applications\label{sec:application}}
In this section, we apply our general results to specific vector-valued learning tasks, including MCC and MLC.
Let $\tilde{\ell}:\rbb\mapsto\rbb$ be convex, decreasing and $(L/2)$-Lipschitz continuous. Popular choices of $\tilde{\ell}$ includes the hinge-loss
$\tilde{\ell}(t)=\max\{0,1-t\}$ and the logistic loss $\tilde{\ell}(t)=\log(1+\exp(-t))$.
We only consider applications to RRM here. It should be mentioned that our results directly apply to SGD.

\subsection{Multi-class classification}
MCC is a classic learning problem which aims to assign a \emph{single} class label to each training example, i.e., $\ycal=[c]$. In this problem setting, the $j$-th component of a model $h:\xcal\mapsto\rbb^c$ measures the likelihood of the output label being $j$. The prediction is made by taking the component with the largest value as the predicted label, i.e., $x\mapsto\arg\max_{j\in[c]}h_j(x)$. We apply our result to three classical MCC models: the multi-class SVM, the multinomial logistic regression and top-k SVM.
In Table \ref{tab:mcc}, we compare generalization bounds for MCC in a Frobenius learning framework, i.e., either there is a constraint $\|W\|_F\leq 1$ or a regularizer $\|W\|_F^2$.

\begin{example}[Multi-class SVM\label{exp:cs}]
  Let $\ycal=[c]$. Consider the objective function $F_S$ in \eqref{F-S} with \[\ell(\bw;z)=\max_{y'\in[c]:y'\neq y}\tilde{\ell}\big(\langle\bw_y-\bw_{y'},\phi(x)\rangle\big)\] and $r(\bw)=\frac{\sigma}{2}\sum_{j\in[c]}\|\bw_j\|_2^2$. This is a margin-based loss function and $\min_{y'\in[c]:y'\neq y}\langle\bw_y-\bw_{y'},x\rangle$ is called the margin of $h^{\bw}$ at $z=(x,y)$. It is clear that the loss function is designed to favor a model with a large margin. This objective function recovers the multi-class SVM in \citet{crammer2002algorithmic} by taking $\tilde{\ell}$ as the hinge loss. To apply Theorem \ref{thm:main}, it suffices to verify Assumptions \ref{ass:lip}, \ref{ass:r}. It was shown that $\ell$ is $L$-Lipschitz continuous w.r.t. $\|\cdot\|_\infty$~\citep{lei2019data}. The convexity of $\ell$ follows from the convexity of $\tilde{\ell}$ and the linearity of hypothesis. The $\sigma$-strong convexity of $r$ w.r.t. $\|\cdot\|_{2,2}$ is clear in the literature~\citep{kakade2012regularization}. The dual norm is $\|\cdot\|_{2,2,}$ and therefore $\widetilde{B}\leq\kappa$. Therefore, we can apply Theorem \ref{thm:main} to develop the generalization bound $F(\bw_S)-F(\bw^*)=\widetilde{O}\Big(\frac{\log(1/\delta)}{n\sigma}\Big)$ with high probability.
\end{example}

\begin{example}[Multinomial logistic regression]
  Let $\ycal=[c]$. Consider the loss function
  \[
  \ell(\bw;z)=\log\Big(\sum_{j\in[c]}\exp\big(\langle\bw_j-\bw_y,\phi(x)\rangle\big)\Big).
  \]
  Consider the objective function $F_S$ \eqref{F-S} with the above loss function and the regularizer $r(\bw)=\frac{\sigma}{2}\sum_{j\in[c]}\|\bw_j\|_2^2$. This learning scheme recovers the popular multinomial logistic regression.
  It was shown that the above loss function is $2$-Lipschitz continuous w.r.t. $\|\cdot\|_\infty$~\citep{lei2019data}.  Assumption \ref{ass:r} also holds.
  Therefore, we can apply Theorem \ref{thm:main} to develop the generalization bound $F(\bw_S)-F(\bw^*)=\widetilde{O}\Big(\frac{\log(1/\delta)}{n\sigma}\Big)$ with probability at least $1-\delta$.
\end{example}

\begin{example}[Top-k SVM]
  Let $\ycal=[c]$. Consider the loss
  \begin{multline*}
  \ell(\bw;z)=\max\Big\{0,\frac{1}{k}\sum_{j=1}^{c}\big(\ibb_{y\neq1}+\langle\bw_1-\bw_y,\phi(x)\rangle,\cdots,\\
  \ibb_{y\neq c}+\langle\bw_c-\bw_y,\phi(x)\rangle\big)_{\{j\}}\Big\},
  \end{multline*}
  where $\ibb$ is the indicator function and for any $\bt\in\rbb^c$ the notation $\{\cdot\}$ denotes a permutation such that $\{j\}$ is the index of the $j$-th largest score, i.e., $t_{\{1\}}\geq t_{\{2\}}\geq\cdots\geq t_{\{c\}}$. If we use the above loss function and the regularizer $r(\bw)=\frac{\sigma}{2}\sum_{j\in[c]}\|\bw_j\|_2^2$ in \eqref{F-S}, we recover the top-k SVM useful to tackle the ambiguity in class labels~\citep{lapin2015top}. It was shown that the above loss function is $2$-Lipschitz continuous w.r.t. $\|\cdot\|_\infty$~\citep{lei2019data}.  Assumption \ref{ass:r} also holds.
  Therefore, we can apply Theorem \ref{thm:main} to develop the generalization bound $F(\bw_S)-F(\bw^*)=\widetilde{O}\Big(\frac{\log(1/\delta)}{n\sigma}\Big)$ with probability at least $1-\delta$.
\end{example}
\begin{remark}
  The existing generalization analysis of MCC considers ERM in a constrained hypothesis space~\citep{lei2019data}. However, RRM is more popular in MCC. For example, the multi-class SVM, multinomial logistic regression and top-k SVM in the above three examples are proposed in a regularization setting~\citep{lapin2015top,crammer2002algorithmic}. Furthermore, the existing discussions generally imply a slow generalization bound $O(1/\sqrt{n})$~\citep{lei2015multi,lei2019data} or require restrictive assumptions such as low-noise condition and capacity assumptions to achieve a fast generalization bound $O(1/n)$~\citep{li2019learning}. As a comparison, our discussion implies a fast bound $O(1/n)$ without these assumptions.
\end{remark}

\subsection{Multi-label classification}
For MLC, each training example can be associated with one or more class labels~\citep{yu2014large,zhang2013review,xu2020partial,dembczynski2012label,wu2020multi}. This can be realized by setting $\ycal=\{-1,+1\}^c$, i.e., each $y=\big(y^{(1)},\ldots,y^{(c)}\big)$ is a binary vector where $y^{(j)}=1$ if the $j$-th label is relevant and $y^{(j)}=-1$ if the $j$-th label is irrelevant. A popular approach to MLC is to learn a vector-valued function $h:\xcal\mapsto\rbb^c$ and predict the output $\hat{y}$ for $z$ according to its sign, i.e., $\hat{y}^{(j)}=\text{sgn}(h_j(x))$, where $\text{sgn}(a)$ denotes the sign of $a\in\rbb$. There are various performance measures to quantify the performance of a multi-label predictor, including the subset accuracy and the ranking loss. In Table \ref{tab:mlc}, we compare generalization bounds for MLC in a Frobenius learning framework, i.e., either a constraint $\|W\|_F\leq 1$ or a regularizer $\|W\|_F^2$.

\begin{example}[Learning with subset loss]
  Let $\ycal=\{-1,+1\}^c$. Consider the loss function
  \begin{equation}\label{subset}
  \ell(\bw;z)=\max_{j\in[c]}\tilde{\ell}\big(y^{(j)}\langle\bw_j,\phi(x)\rangle\big).
  \end{equation}
  This loss function is called the subset loss~\citep{zhang2013review}. Note that $\tilde{\ell}(y^{(j)}\langle\bw_j,\phi(x)\rangle)$ is a standard loss if we consider the prediction of the $j$-th label as a standard binary classification problem. Then this loss function encourages us to predict all labels correctly. As we will show in Proposition \ref{prop:loss-label}, the subset loss is $L$-Lipschitz continuous w.r.t. $\|\cdot\|_\infty$. If we consider the objective function \eqref{F-S} with the subset loss and the regularizer $r(\bw)=\frac{\sigma}{2}\sum_{j\in[c]}\|\bw_j\|_2^2$, then we can apply Theorem \ref{thm:main} to derive with probability $1-\delta$ that $F(\bw_S)-F(\bw^*)=\widetilde{O}\Big(\frac{\log(1/\delta)}{n\sigma}\Big)$.
\end{example}

\begin{example}[Learning with ranking loss]
  Let $\ycal=\{-1,+1\}^c$. For each $y\in\{-1,+1\}^c$ we denote $y_+=\{j\in[c]:y^{(j)}=+1\}$ and $y_-=\{j\in[c]:y^{(j)}=-1\}$ as the set of relevant and irrelevant labels, respectively. Consider the following ranking loss~\citep{zhang2013review}
  \begin{equation}\label{rank-loss}
    \ell(\bw;z)=\frac{1}{|y_+||y_-|}\sum_{j_+\in y_+}\sum_{j_-\in y_-}\tilde{\ell}\big(\langle\bw_{j_+}-\bw_{j_-},\phi(x)\rangle\big),
  \end{equation}
  where $|A|$ denotes the cardinality of a set $A$. Intuitively, the ranking loss encourages predictors with larger function values for a relevant label than a irrelevant label. As we will show in Proposition \ref{prop:loss-label}, the ranking loss is $L$-Lipschitz continuous w.r.t. $\|\cdot\|_\infty$. If we consider the objective function \eqref{F-S} with the ranking loss and the regularizer $r(\bw)=\frac{\sigma}{2}\sum_{j\in[c]}\|\bw_j\|_2^2$, then we can apply Theorem \ref{thm:main} to show with probability at least $1-\delta$ that $F(\bw_S)-F(\bw^*)=\widetilde{O}\Big(\frac{\log(1/\delta)}{n\sigma}\Big)$.
\end{example}

\begin{remark}
  Generalization bounds of the order $O(1/\sqrt{n})$ were established for MLC with the decomposable loss $\ell(\bw;z)=\frac{1}{c}\sum_{j\in[c]}\big(y^{(j)}-\langle\bw_j,x\rangle\big)^2$ under ERM in a constrained space $\{\bw\in\rbb^{d\times c}:\|\bw\|_{S_1}\leq\Lambda\}$~\citep{yu2014large}. Although this bound has no dependency on $c$, it requires a constraint in terms of $\|\cdot\|_{S_1}$, which can be as large as $\sqrt{c}\|\cdot\|_{2,2}$ (i.e., a constraint $\|\bw\|_{S_1}\leq\Lambda$ corresponds to the constraint $\|\cdot\|_{2,2,}\leq\sqrt{c}\Lambda$). As a comparison, we consider a regularization scheme involving the much milder norm $\|\cdot\|_{2,2}$. Also, there is a gap between the theoretical analysis and the algorithm design: the generalization analysis there is for ERM, while the algorithm is designed based on a regularization scheme~\citep{yu2014large}. Furthermore, our analysis for regularization implies a fast rate $O(1/n)$.
\end{remark}

%
\begin{figure*}[h!]
  \setlength{\abovecaptionskip}{0pt}
  \setlength{\belowcaptionskip}{6pt}
  \centering
  \subfigure[ALOI]{\includegraphics[width=0.25\textwidth, trim=0 0 0 0, clip]{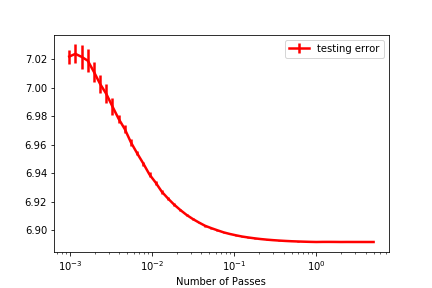}}\hspace*{-0.29cm} 
  \subfigure[CIFAR10]{\includegraphics[width=0.25\textwidth, trim=0 0 0 0, clip]{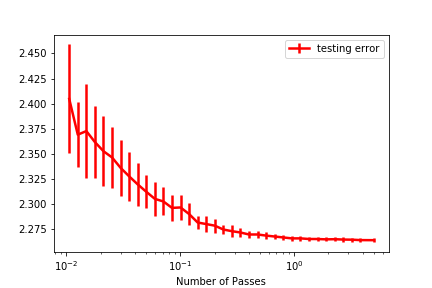}}\hspace*{-0.29cm}
  \subfigure[RCV1]{\includegraphics[width=0.25\textwidth, trim=0 0 0 0, clip]{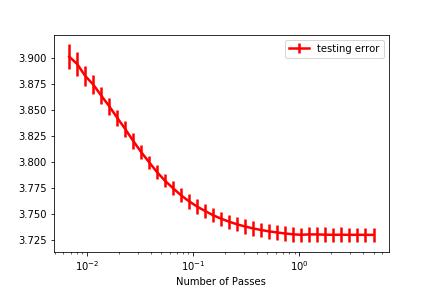}}\hspace*{-0.29cm}
  \subfigure[SECTOR]{\includegraphics[width=0.25\textwidth, trim=0 0 0 0, clip]{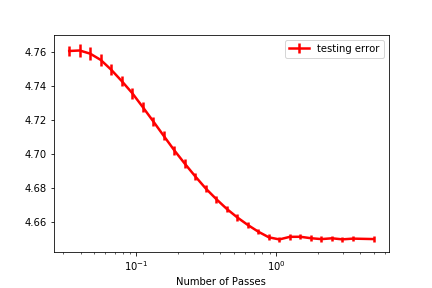}}\hspace*{-0.29cm}
  \vspace*{-0.1\baselineskip}
  \caption{Testing errors versus the number of passes.
  \label{fig:pass}}
  \vspace*{-0.36cm}
\end{figure*}

\begin{figure*}[h]
  \setlength{\abovecaptionskip}{0pt}
  \setlength{\belowcaptionskip}{6pt}
  \centering
  \subfigure[ALOI]{\includegraphics[width=0.25\textwidth, trim=0 0 0 0, clip]{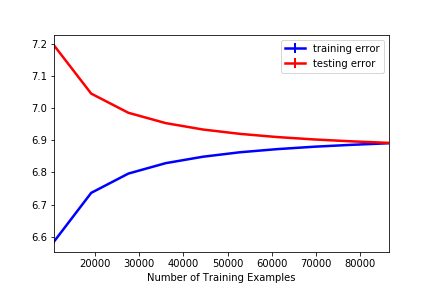}}\hspace*{-0.29cm} 
  \subfigure[CIFAR10]{\includegraphics[width=0.25\textwidth, trim=0 0 0 0, clip]{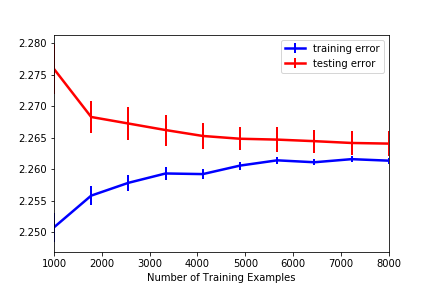}}\hspace*{-0.29cm}
  \subfigure[RCV1]{\includegraphics[width=0.25\textwidth, trim=0 0 0 0, clip]{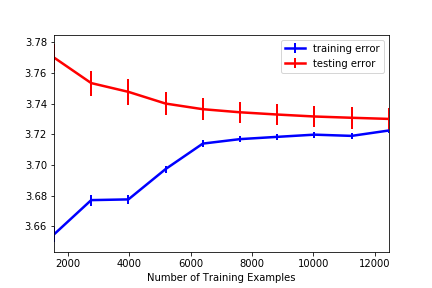}}\hspace*{-0.29cm}
  \subfigure[SECTOR]{\includegraphics[width=0.25\textwidth, trim=0 0 0 0, clip]{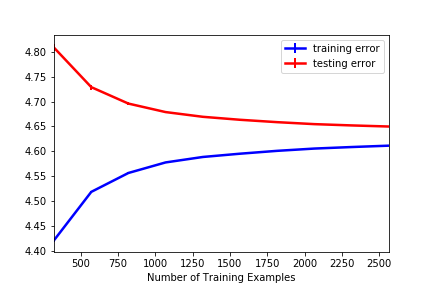}}\hspace*{-0.29cm}
  \vspace*{-0.1\baselineskip}
  \caption{Training and testing errors versus the training data size.
  \label{fig:sample-size}}
  \vspace*{-0.36cm}
\end{figure*}
\begin{figure*}[h!]
  \setlength{\abovecaptionskip}{0pt}
  \setlength{\belowcaptionskip}{6pt}
  \centering
  \subfigure[ALOI]{\includegraphics[width=0.25\textwidth, trim=0 0 0 0, clip]{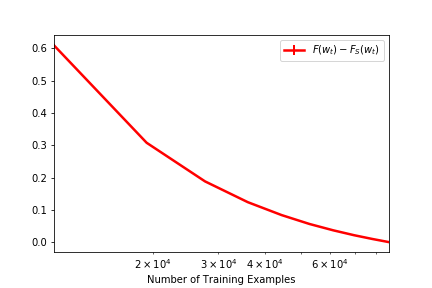}}\hspace*{-0.29cm} 
  \subfigure[CIFAR10]{\includegraphics[width=0.25\textwidth, trim=0 0 0 0, clip]{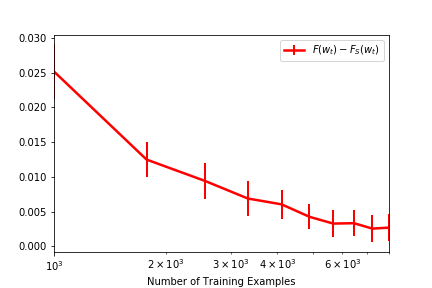}}\hspace*{-0.29cm}
  \subfigure[RCV1]{\includegraphics[width=0.25\textwidth, trim=0 0 0 0, clip]{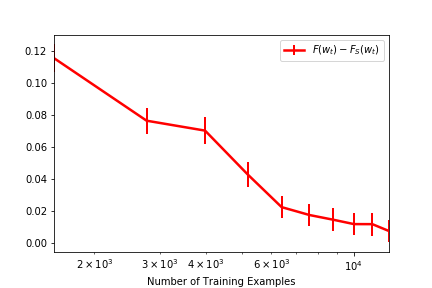}}\hspace*{-0.29cm}
  \subfigure[SECTOR]{\includegraphics[width=0.25\textwidth, trim=0 0 0 0, clip]{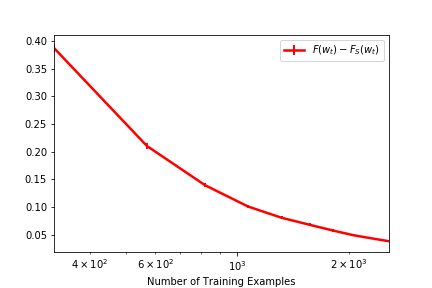}}\hspace*{-0.29cm}
  \vspace*{-0.1\baselineskip}
  \caption{$F(\bw_T)-F_S(\bw_T)$ versus the training data size.
  \label{fig:diff}}
\vspace{-0.2cm}
\end{figure*}
The proof of Proposition \ref{prop:loss-label} is given in Appendix \ref{sec:loss-label}.
\begin{proposition}\label{prop:loss-label}
\begin{enumerate}
  \item If $\tilde{\ell}$ is $L$-Lipschitz continuous, then the subset loss \eqref{subset} is $L$-Lipschitz continuous w.r.t. $\|\cdot\|_\infty$.
  \item If $\tilde{\ell}$ is $(L/2)$-Lipschitz continuous, then the ranking loss \eqref{rank-loss} is $L$-Lipschitz continuous w.r.t. $\|\cdot\|_\infty$.
\end{enumerate}
\end{proposition}

\section{Experimental Verification\label{sec:experiment}}

In this section, we present experimental results to verify our theoretical analysis. We consider a specific vector-valued learning problem called multinomial logistic regression, where the aim is to predict a class label for each training example. We apply SGD \eqref{SGD} to solve the optimization problem with the objective function
\begin{equation}\label{MLG}
  F_S(\bw)=
  \frac{1}{n}\sum_{i=1}^{n}\log\Big(\sum_{j\in[c]}\exp\big(\langle\bw_j-\bw_{y_i},x_i\rangle\big)\Big)+\frac{\lambda}{2}\|\bw\|_{2,2}^2,
\end{equation}
where $\bw=(\bw_1,\ldots,\bw_c)\in\rbb^{d\times c}$ and $\lambda=0.01$. We set the initial point $\bw=0$ and the step size $\eta_t=1/(\lambda t+1)$. We consider four real-world datasets available from the LIBSVM homepage~\citep{chang2011libsvm}, whose information is summarized in Table \ref{tab:data_set}.
We repeat experiments 50 times and report the average as well as standard deviation of the experimental results.
We call $F_S$ and $F$ the training error and testing error, respectively.
We consider two experiments.

For the first experiment, we aim to verify the generalization bound in \eqref{thm-sgd-res}. We randomly use 80\% of data for training and reserve the remaining 20\% for testing. We plot the testing errors $F(\bw_t)$ of the SGD sequence versus the number of passes, i.e., $t/n$ in Figure \ref{fig:pass}. It is clear that the testing error decreases initially as we run more SGD iterations. After sufficient number of iterations, the generalization performance of SGD iterates no long improves. This is well consistent to the generalization bound in \eqref{thm-sgd-res} which shows that the generalization bound would be dominated by the effect due to sample size if $T\geq n$.

For the second experiment, our aim is to show whether the dependency of the generalization bound \eqref{thm-sgd-res-b} on the sample size can be really captured in practice. For each dataset, we randomly select different number of examples for training and reserve the remaining for testing. For each fixed number of training examples, we train a model by solving \eqref{MLG} with SGD and stop it after $5$ passes over the data.
We compute both the training error $F_S(\bw_T)$ and testing error $F(\bw_T)$ of the output model $\bw_T$ (last iterate).
In this way we get a training error and a testing error for each considered sample size.
We then plot the relative behavior of these errors versus the number of training examples in Figure \ref{fig:sample-size}. According to Figure \ref{fig:sample-size}, it is clear that training errors increase as the increase of training examples. This phenomenon is due to the increasing difficulty of the optimization problem as the increase of training examples. Note the fit of a model on $10,000$ examples would be significantly harder than that on $100$ examples. It is also clear that the generalization behavior improves as we have more training examples. This matches well the generalization bound \eqref{thm-sgd-res-b}. In Figure \ref{fig:diff} we further plot the difference between testing error and training error ($F(\bw_T)-F_S(\bw_T)$) versus the sample size. It is clear this difference is a decreasing function of the sample size.

\section{Conclusion\label{sec:conclusion}}
We present a unifying generalization analysis for the regularization framework of learning with vector-valued functions. Our generalization bounds admit a mild dependency on the output dimension and decay fast w.r.t. the sample size.
For instance for MLC, they improve the best known dependency on the number of classes from $O(\sqrt{c})$ to a $O(\log(c))$, making them suitable for extreme classification.
Furthermore, our analysis  relax the existing restrictive assumptions such as
a low noise condition or a smoothness requirement on loss functions. We develop the first generalization analysis for SGD to learn vector-valued functions, which is important to understand both statistical properties of models and the convergence of the algorithm. We present applications to specific learning machines and conduct experiments to verify our theory.

It would be interesting to extend our discussion to a distributed learning setting~\citep{lin2018distributed,hu2020distributed}. It would be also interesting to study non-strongly convex regularizers such as $\ell_1$ regularizers which is useful to learn sparse models~\citep{guo2017thresholded}.


\onecolumn
\appendix
\numberwithin{equation}{section}
\numberwithin{theorem}{section}
\numberwithin{figure}{section}
\numberwithin{table}{section}
\renewcommand{\thesection}{{\Alph{section}}}
\renewcommand{\thesubsection}{\Alph{section}.\arabic{subsection}}
\renewcommand{\thesubsubsection}{\Roman{section}.\arabic{subsection}.\arabic{subsubsection}}
\setcounter{secnumdepth}{-1}
\setcounter{secnumdepth}{3}

\section{Proofs\label{sec:proof}}
\subsection{Proof of Theorem \ref{thm:rademacher}\label{sec:proof-rademacher}}
The following lemma relates the Rademacher complexity of loss function classes to that of hypothesis spaces~\citep{lei2019data}. The idea of considering a function space $\widetilde{\hcal}_\Lambda$ defined over an extended input argument is to fully exploit the Lipschitz continuity of loss functions w.r.t. the infinity norm, i.e.,
\[
\max_z\big|\ell(\bw;z)-\ell(\bw';z)\big|\leq L\max_x\max_j\big|h^{\bw_j}(x)-h^{\bw_j'}(x)\big|.
\]
We define the \emph{worst-case} Rademacher complexity as $\mathfrak{R}_n(\hcal)=\sup_{S\in{\zcal}^n}\mathfrak{R}_{S}(\hcal)$.
\begin{lemma}[\label{lem:lei2019}\citealtt{lei2019data}]
Let Assumption \ref{ass:lip} hold. For any $\Lambda>0$, define
\begin{equation}
\label{tilde-H}\widetilde{\hcal}_{\Lambda}=\Big\{(x,j)\mapsto h^{\bw_j}(x):x\in\xcal,j\in[c],\bw\in\wcal_\Lambda\Big\}.
\end{equation}
Then there exists a constant $C_1$ independent of $n,c,L$ and $\Lambda$ such that
\[\mathfrak{R}_S(\fcal_\Lambda)\leq C_1L\sqrt{c}\mathfrak{R}_{nc}\big(\widetilde{\hcal}_\Lambda\big)\log^{2}(nc).\]
\end{lemma}
To apply the above lemma, we need to estimate $\mathfrak{R}_{nc}\big(\widetilde{\hcal}_\Lambda\big)$. This is achieved in the following lemma. Note that the upper and lower bound matches up to a constant factor.
Note $\widetilde{\hcal}_{\Lambda}$ takes the form  $\widetilde{\hcal}_{\Lambda}=\big\{(x,j)\mapsto\langle\bw_j,x\rangle:\bw\in\wcal_\Lambda\big\}$ in this paper. Recall that $\bp_j(x):=\big(\underbrace{0,\ldots,0}_{j-1},\phi(x),\underbrace{0,\ldots,0}_{c-j}\big)\in \hcal_K^c.$
\begin{lemma}\label{lem:rademacher}
If Assumption \ref{ass:r} holds,
then the worst-case Rademacher complexity of $\widetilde{\hcal}_\Lambda$ can be bounded by
\[
\sqrt{\frac{1}{2nc}}\sup_{\bw\in\wcal_\Lambda}\big\langle\bw,\bp_{j}(x)\big\rangle\leq\mathfrak{R}_{nc}(\widetilde{\hcal}_\Lambda)\leq \sqrt{\frac{2\Lambda}{nc\sigma}}\widetilde{B}.
\]
\end{lemma}
For a convex function $f$, we denote by $f^*$ its Fenchel conjugate, i.e., $f^*(\bv):=\sup_{\bw}[\langle \bw,\bv\rangle-f(\bw)]$. The following lemma is due to
\citet{kakade2012regularization}.
\begin{lemma}\label{lem:conjugate}
  If $\tau$ is $\sigma$-strongly convex w.r.t. $\|\cdot\|$ and $\tau^*(0)=0$, then for any sequence $\bv_1,\ldots,\bv_n$ and any $\bw$ we have
  \[
  \sum_{i=1}^{n}\langle\bw,\bv_i\rangle\leq \tau(\bw)+\sum_{i=1}^{n}\langle\nabla \tau^*(\bv_{1:i-1}),\bv_i\rangle+\frac{1}{2\sigma}\sum_{i=1}^n\|\bv_i\|_*^2,
  \]
  where $\bv_{1:i-1}$ denotes the sum $\sum_{j=1}^{i-1}\bv_j$.
\end{lemma}
\begin{proof}[Proof of Lemma \ref{lem:rademacher}]
Define $\tau:\wcal\mapsto\rbb$ by $\tau(\bw)=F(\bw)-F(\bw^*)$. It is clear that $\tau$ is $\sigma$-strongly convex w.r.t. $\|\cdot\|$. Furthermore, it follows from the definition of $\bw^*$ that
\[
\tau^*(0)=\sup_{\bv}\Big(\langle0,\bv\rangle-\tau(\bv)\Big)=-\inf_{\bv}\tau(\bv)=0.
\]
According to the definition of Rademacher complexity~\citep{bartlett2002rademacher} and $\widetilde{\hcal}_\Lambda$ in \eqref{tilde-H}, we know
\begin{equation}\label{rademacher-1}
nc\mathfrak{R}_{nc}(\widetilde{\hcal}_\Lambda)=\sup_{(x_i,j_i):i\in[n]}\ebb_\epsilon\sup_{\bw\in\wcal_\Lambda}\sum_{i=1}^{nc}\epsilon_i\langle\bw_{j_i},\phi(x_i)\rangle
=\sup_{(x_i,j_i):i\in[n]}\ebb_\epsilon\sup_{\bw\in\wcal_\Lambda}\sum_{i=1}^{nc}\epsilon_i\langle\bw,\tilde{\phi}_{j_i}(x_i)\rangle,
\end{equation}
where we have used the definition of $\bp_{j_i}$ in the last identity.
We apply Lemma \ref{lem:conjugate} with $\bv_i=\lambda\epsilon_i\bp_{j_i}(x_i)$ and derive
\[
\sum_{i=1}^{nc}\epsilon_i\langle\bw,\lambda\bp_{j_i}(x_i)\rangle\leq\tau(\bw)+\sum_{i=1}^{nc}\langle\nabla\tau^*(\bv_{1:i-1}),\bv_i\rangle+\frac{\lambda^2}{2\sigma}
\sum_{i=1}^{nc}\|\bp_{j_i}(x_i)\|_*^2.
\]
It is clear that
\[
\ebb_\epsilon\langle\nabla\tau^*(\bv_{1:i-1}),\bv_i\rangle=\ebb_\epsilon\Big\langle\nabla\tau^*\Big(\lambda\sum_{k=1}^{i-1}\epsilon_k\bp_{j_k}(x_k)\Big),
\lambda\epsilon_i\bp_{j_i}(x_i)\Big\rangle=0
\]
and therefore
\[
\ebb_\epsilon\sup_{\bw\in\wcal_\Lambda}\sum_{i=1}^{nc}\epsilon_i\langle\bw,\bp_{j_i}(x_i)\rangle\leq\frac{\sup_{\bw\in\wcal_\Lambda}\tau(\bw)}{\lambda}+\frac{\lambda}{2\sigma}\sum_{i=1}^{nc}
\|\bp_{j_i}(x_i)\|_*^2.
\]
We can combine the above inequality and \eqref{rademacher-1} together and get
\begin{align*}
  nc\mathfrak{R}_{nc}(\widetilde{\hcal}_\Lambda) & \leq \frac{\Lambda}{\lambda}+\frac{\lambda}{2\sigma}\sup_{(x_i,j_i):i\in[n]}\sum_{i=1}^{nc}\|\bp_{j_i}(x_i)\|_*^2
  = \frac{\Lambda}{\lambda}+\frac{\lambda nc}{2\sigma}\widetilde{B}^2.
\end{align*}
With the choice $\lambda=\Big(\frac{2\Lambda\sigma}{nc\widetilde{B}^2}\Big)^{\frac{1}{2}}$, we derive
\[
nc\mathfrak{R}_{nc}(\widetilde{\hcal}_\Lambda)\leq \sqrt{2\Lambda nc\sigma^{-1}}\widetilde{B}.
\]
This establishes the upper bounds.
We now turn to the lower bounds. By taking $x_1=\cdots=x_n=x,j_1=\cdots=j_n=j$, we know ($\wcal_\Lambda$ is symmetric)
\[
\sup_{\bw\in\wcal_\Lambda}\Big\langle\bw,\sum_{i=1}^{nc}\epsilon_i\bp_{j_i}(x_i)\Big\rangle
= \sup_{\bw\in\wcal_\Lambda}\Big\langle\bw,\sum_{i=1}^{nc}\epsilon_i\bp_{j}(x)\Big\rangle
=\Big|\sum_{i=1}^{nc}\epsilon_i\Big|\sup_{\bw\in\wcal_\Lambda}\big\langle\bw,\bp_{j}(x)\big\rangle.
\]
It then follows from \eqref{rademacher-1} and Khitchine's inequality $\ebb_\epsilon|\sum_{i=1}^{n}\epsilon_i|\geq \sqrt{n/2}$~\citep{mohri2012foundations} that
\begin{align*}
\sup_{(x_i,j_i):i\in[n]}\ebb_\epsilon\sup_{\bw\in\wcal_\Lambda}\sum_{i=1}^{nc}\epsilon_i\langle\bw_{j_i},\phi(x_i)\rangle
& \geq \sup_{(x,j)}\ebb_\epsilon\Big|\sum_{i=1}^{nc}\epsilon_i\Big|\sup_{\bw\in\wcal_\Lambda}\big\langle\bw,\bp_{j}(x)\big\rangle\\
& \geq \sqrt{nc/2}\sup_{(x,j)}\sup_{\bw\in\wcal_\Lambda}\big\langle\bw,\bp_{j}(x)\big\rangle.
\end{align*}
This establishes the stated lower bound and finishes the proof.
\end{proof}
\begin{proof}[Proof of Theorem \ref{thm:rademacher}]
  Theorem follows directly by combining Lemma \ref{lem:lei2019} and Lemma \ref{lem:rademacher} together.
\end{proof}

\subsection{Proof of Theorem \ref{thm:main}\label{sec:proof-main}}
To prove Theorem \ref{thm:main}, we first introduce some necessary notations.
Let $\rho_0>0$ and $\delta_0\in(0,1)$ be two numbers to be fixed. We construct two sequences
$\rho_k=2^k\rho_0,\delta_k=2^{-k}\delta_0$ for $k\in\nbb$. For brevity we set $\rho_{-1}=0$.
We decompose $\wcal$ into a sequence of disjoint sets according to the function value
\[
\wcal_k'=\Big\{\bw\in\wcal:\rho_{k-1}<F(\bw)-F(\bw^*)\leq\rho_k\Big\}.
\]
For brevity, for any $f:\wcal\times\zcal\mapsto\rbb$ we denote $\ebb[f(\bw;\cdot)]$ the expectation $\ebb_z[f(\bw;z)]$ and $\hat{\ebb}_S[f(\bw;\cdot)]=\frac{1}{n}\sum_{i=1}^{n}f(\bw;z_i)$ the empirical average.
For any $\bw$, we define $s(\bw,\cdot)$ the shifted loss relative to $\bw^*$
\[
s(\bw;z)=\ell(\bw;z)-\ell(\bw^*;z).
\]

Our proof of Theorem \ref{thm:main} requires to use a concentration inequality called McDiarmid's inequality~\citep{mohri2012foundations}.
\begin{lemma}[McDiarmid's inequality]\label{lem:mcdiarmid}
  Let $Z_1,\ldots,Z_n$ be independent random variables taking values in a set $\zcal$, and assume that $f:\zcal^n\mapsto\rbb$ satisfies
  \begin{equation}\label{bounded-variation-assumption}
    \sup_{z_1,\ldots,z_n,\bar{z}_i\in\zcal}|f(z_1,\cdots,z_n)-
    f(z_1,\cdots,z_{i-1},\bar{z}_i, z_{i+1},\cdots,z_n)| \leq c_i
  \end{equation}
  for $1\leq i\leq n$. Then, for any $0<\delta<1$, with probability of at least $1-\delta$, we have
  $$f(Z_1,\ldots,Z_n)\leq\ebb f(Z_1,\ldots,Z_n)+\sqrt{\frac{\sum_{i=1}^nc_i^2\log(1/\delta)}{2}}.$$
\end{lemma}
\begin{proof}[Proof of Theorem \ref{thm:main}]
By the $\sigma$-strong convexity of $F$, we know
$F(\bw)-F(\bw^*)\geq\frac{\sigma}{2}\|\bw-\bw^*\|^2$,
from which we derive the following inequality for any $\bw\in\wcal_k'$
\[
\|\bw-\bw^*\|\leq \Big(2\sigma^{-1}\big(F(\bw)-F(\bw^*)\big)\Big)^{\frac{1}{2}}\leq(2\sigma^{-1}\rho_k)^{\frac{1}{2}}.
\]
For any $\bw\in\wcal_k'$ and $z=(x,y)$ we know
\begin{align*}
  |s(\bw;z)| & = \big|\ell(\bw;z)-\ell(\bw^*;z)\big|
    \leq L\big\|h^{\bw}(x)-h^{\bw^*}(x)\big\|_\infty\\
    &= L\Big\|\Big(\langle\bw_1-\bw^*_1,\phi(x)\rangle,\ldots,\langle\bw_c-\bw^*_c,\phi(x)\rangle\Big)\Big\|_\infty
   = L\max_{j=1,\ldots,c}\big|\langle\bw_j-\bw^*_j,\phi(x)\rangle\big| \\
   & \leq L\max_{j=1,\ldots,c}\|\bw_j-\bw_j^*\|_2\|\phi(x)\|_2
   \leq L\kappa\|\bw-\bw^*\|_{2,\infty}\leq L\kappa(2\rho_k/\sigma)^{\frac{1}{2}},
\end{align*}
where we have used the mild assumption $\|\cdot\|_{2,\infty}\leq\|\cdot\|$.

For any $S=\{z_1,\ldots,z_{k-1},z_k,z_{k+1},\ldots,z_n\}$ and $S'=\{z_1,\ldots,z_{k-1},z'_k,z_{k+1},\ldots,z_n\}$, we have
\begin{align*}
   & \Big|\sup_{\bw\in\wcal_k'}\big(\ebb[s(\bw;\cdot)]-\hat{\ebb}_S[s(\bw;\cdot)]\big)-\sup_{\bw\in\wcal_k'}\big(\ebb[s(\bw;\cdot)]-\hat{\ebb}_{S'}[s(\bw;\cdot)]\big)\Big| \\
   & \leq \sup_{\bw\in\wcal_k'}\Big|\hat{\ebb}_S[s(\bw;\cdot)]-\hat{\ebb}_{S'}[s(\bw;\cdot)]\Big| = \frac{1}{n}\sup_{\bw\in\wcal_k'}\big|s(\bw;z_k)-s(\bw;z_k')\big| \\
   & \leq \frac{2}{n}\sup_{\bw\in\wcal_k'}\sup_{z\in\zcal}|s(\bw;z)|\leq \frac{2L\kappa}{n}\big(2\rho_k/\sigma\big)^{\frac{1}{2}}.
\end{align*}

By the McDiarmid's inequality with increments bounded by $\frac{2L\kappa}{n}\big(2\rho_k/\sigma\big)^{\frac{1}{2}}$, we get the following inequality with probability at least $1-\delta_k$
\begin{equation}\label{main-1}
\sup_{\bw\in\wcal_k'}\Big(\ebb[s(\bw;\cdot)]-\hat{\ebb}_S[s(\bw;\cdot)]\Big)
\leq \ebb_S\sup_{\bw\in\wcal_k'}\Big(\ebb[s(\bw;\cdot)]-\hat{\ebb}_S[s(\bw;\cdot)]\Big)+2L\kappa\Big(\frac{\rho_k\log(1/\delta_k)}{n\sigma}\Big)^{\frac{1}{2}}.
\end{equation}

It follows from the symmetry trick that ($\widetilde{S}=\{\tilde{z}_1,\ldots,\tilde{z}_n\}$ is drawn i.i.d. and independently of $S$)
\begin{align}
   &\ebb_S\sup_{\bw\in\wcal_k'}\Big(\ebb[s(\bw;\cdot)]-\hat{\ebb}_S[s(\bw;\cdot)]\Big)  = \ebb_S\sup_{\bw\in\wcal_k'}\Big(\ebb_{\widetilde{S}}\hat{\ebb}_{\widetilde{S}}[s(\bw;\cdot)]-\hat{\ebb}_S[s(\bw;\cdot)]\Big) \notag\\
   & \leq \ebb_{S,\widetilde{S}}\sup_{\bw\in\wcal_k'}\Big(\hat{\ebb}_{\widetilde{S}}[s(\bw;\cdot)]-\hat{\ebb}_S[s(\bw;\cdot)]\Big) = \frac{1}{n}\ebb_{S,\widetilde{S},\epsilon}\sup_{\bw\in\wcal_k'}\Big[\sum_{i=1}^{n}\epsilon_i\big(s(\bw;z_i)-s(\bw;\tilde{z}_i)\big)\Big]\notag\\
   & \leq \frac{2}{n}\ebb_{S,\epsilon}\sup_{\bw\in\wcal_k'}\Big[\sum_{i=1}^{n}\epsilon_is(\bw;z_i)\Big]= \frac{2}{n}\ebb_{S,\epsilon}\sup_{\bw\in\wcal_k'}\sum_{i=1}^{n}\epsilon_i\ell(\bw;z_i)=2\ebb_S\mathfrak{R}_S(\fcal_{\rho_k}),\label{main-2-1}
\end{align}
where in the last second step we remove $\ell(\bw^*;z_i)$ since $\bw^*$ is fixed.
According to Lemma \ref{lem:lei2019} and Lemma \ref{lem:rademacher}, we know
\begin{align*}
  \ebb_S\mathfrak{R}_S(\fcal_{\rho_k}) & \leq C_1L\sqrt{c}\mathfrak{R}_{nc}\big(\widetilde{\hcal}_{\rho_k}\big)\log^2(nc)
  \leq \frac{\sqrt{2\rho_k}C_1L\log^2(nc)\widetilde{B}}{\sqrt{n\sigma}}.
\end{align*}
Combining \eqref{main-1}, \eqref{main-2-1} and the above inequality together, we derive the following inequality with probability at least $1-\delta_k$
\begin{equation}\label{main-2}
\sup_{\bw\in\wcal_k'}\Big(\ebb[s(\bw;\cdot)]-\hat{\ebb}_S[s(\bw;\cdot)]\Big)
\leq \frac{2\sqrt{2\rho_k}C_1L\log^2(nc)\widetilde{B}}{\sqrt{n\sigma}}+2L\kappa\Big(\frac{\rho_k\log(1/\delta_k)}{n\sigma}\Big)^{\frac{1}{2}}.
\end{equation}
For any $\bw\in\wcal_k'$, we have
\[
\frac{\rho_k}{2}=\rho_{k-1}\leq F(\bw)-F(\bw^*)
\]
and therefore
\[
\rho_k\leq\max\big\{\rho_0,2(F(\bw)-F(\bw^*))\big\}.
\]
Furthermore, it is clear
\[
\frac{1}{\delta_k}=\frac{\rho_02^k}{\delta_0\rho_0}\leq\frac{\max\big\{\rho_0,2(F(\bw)-F(\bw^*))\big\}}{\delta_0\rho_0}.
\]
By the definition of $F$, $F_S$ and $s(\bw)$, we know
\[
\big(F(\bw)-F(\bw^*)\big)-\big(F_S(\bw)-F_S(\bw^*)\big)=\ebb[s(\bw,\cdot)]-\hat{\ebb}_S[s(\bw,\cdot)].
\]
Combining the above discussions and \eqref{main-2} together, with probability $1-\delta_k$ the following inequality holds uniformly for all $\bw\in\wcal_k'$
\begin{multline}\label{main-2-2}
  F(\bw)-F(\bw^*) \leq  \big(F_S(\bw)-F_S(\bw^*)\big) +
  2L\sqrt{\frac{\max\big\{\rho_0,2(F(\bw)-F(\bw^*))\big\}}{n\sigma}}\times\\ \Big(\sqrt{2}C_1\log^2(nc)\widetilde{B}+
  \kappa\sqrt{\log(1/\delta_0)+\log\max\big\{1,2\rho_0^{-1}(F(\bw)-F(\bw^*))\big\}}\Big).
\end{multline}
Noticing $\sum_{k=0}^{\infty}\delta_k=2\delta_0$, the above inequality holds with probability at least $1-2\delta_0$ uniformly for all $\bw\in\wcal$.
The stated inequality \eqref{main-a} then follows by setting $\rho_0=1/(n\sigma)$.

We now prove \eqref{main-b}. We prove instead
\begin{equation}\label{main-4}
F(\bw_S)-F(\bw^*)\leq \frac{16L^2}{n\sigma}\Big(2C_1^2\log^{3}(nc)\widetilde{B}^2+4\kappa^2\log(1/\delta_0)\Big).
\end{equation}
Since $F_S(\bw_S)\leq F_S(\bw^*)$, it follows from \eqref{main-2-2} that the following inequality holds with probability $1-2\delta_0$
\begin{multline}\label{main-3}
  F(\bw_S)-F(\bw^*) \leq
  2L\sqrt{\frac{\max\big\{\rho_0,2(F(\bw_S)-F(\bw^*))\big\}}{n\sigma}}\times\\ \Big(\sqrt{2}C_1\log^2(nc)\widetilde{B}+
  \kappa\sqrt{\log(1/\delta_0)+\log\max\big\{1,2\rho_0^{-1}(F(\bw_S)-F(\bw^*))\big\}}\Big).
\end{multline}
Note that the above inequality holds for any $\rho_0>0$. We can take $\rho_0=64L^2\kappa^2/(n\sigma)$.
We now consider two cases. For the case $F(\bw_S)-F(\bw^*))<\rho_0/2$, the inequality \eqref{main-4} is trivial. We now consider the case $F(\bw_S)-F(\bw^*))\geq\rho_0/2$. In this case,
it follows from \eqref{main-3} and the elementary inequality $(a+b)^2\leq2(a^2+b^2)$ that
\[
F(\bw_S)-F(\bw^*)\leq \frac{16L^2}{n\sigma}\Big(2C_1^2\log^{3}(nc)\widetilde{B}^2+
  \kappa^2\Big(\log(1/\delta_0)+\log\big(2\rho_0^{-1}(F(\bw_S)-F(\bw^*))\big)\Big)\Big).
\]
By the elementary inequality $\log a \leq ab+\log(1/b)-1$ for all $a,b>0$, we know
\[
\log\big(2\rho_0^{-1}(F(\bw_S)-F(\bw^*))\big)\leq \frac{n\sigma}{32L^2\kappa^2}(F(\bw_S)-F(\bw^*))+\log(64L^2\kappa^2/(\rho_0n\sigma))-1.
\]
It then follows that
\[
F(\bw_S)-F(\bw^*)\leq \frac{16L^2}{n\sigma}\Big(2C_1^2\log^{3}(nc)\widetilde{B}^2+\kappa^2\log(1/\delta_0)\Big)
+\frac{F(\bw_S)-F(\bw^*)}{2}+\frac{16L^2\kappa^2\log(64L^2\kappa^2/(\rho_0n\sigma))}{n\sigma}.
\]
Since $\rho_0=64L^2\kappa^2/(n\sigma)$ we then get the stated inequality \eqref{main-4}. The proof is complete.
\end{proof}

\subsection{Proof of Theorem \ref{thm:sgd}\label{sec:proof-sgd}}
To prove Theorem \ref{thm:sgd}, we first introduce a high-probability bound on the convergence rate of SGD.
\begin{lemma}[\label{lem:sgd}\citealtt{harvey2018tight}]
  Suppose $F_S(\bw)=\frac{1}{n}\sum_{i=1}^{n}f(\bw;z)$ is $\sigma$-strongly convex w.r.t. $\|\cdot\|_{2,2}$. Assume for all $\bw\in\wcal$ and $z\in\zcal$ there holds $\|f'(\bw;z)\|_{2,2,}\leq \widetilde{L}$ for some $\widetilde{L}>0$. Let $\{\bw_t\}$ be obtained by $\bw_{t+1}=\bw_t-\eta_tf'(\bw_t;z_{i_t})$, where $\eta_t=1/(\sigma t)$ and $\{i_t\}_{t\in\nbb}$ is independently drawn from the uniform distribution over $[n]$. Then with probability at least $1-\delta$ we have \[F_S(\bw_{T})-\inf_{\bw}F_S(\bw)=O\Big(\frac{\log(T)\log(1/\delta)}{T\sigma}\Big).\]
\end{lemma}
\begin{proof}[Proof of Theorem \ref{thm:sgd}]
  For any $j\in[c]$ denote $\ell^{(j)}(\bw;z)=\frac{\partial\ell(\bw;z)}{\partial\bw_j}$ the partial (sub)gradient w.r.t. the $\bw_j$. Then $\ell^{(j)}(\bw;z)=\psi^{(j)}(h^{\bw}(x),y)\phi(x)$, where $\psi^{(j)}$ denotes the partial (sub)gradient w.r.t. the $j$-th coordinate. Assumption \ref{ass:lip} means that $\big\|\big(\psi^{(j)}(h^{\bw}(x),y)\big)_{j=1}^c\big\|_1\leq L$ for all $x,y$. It then follows that
  \begin{align}
    \|\ell'(\bw;z)\|_{2,2}^2 & = \sum_{j=1}^{c}\big\|\ell^{(j)}(\bw;z)\big\|_2^2 = \sum_{j=1}^{c}\big|\psi^{(j)}(h^{\bw}(x),y)\big|^2\|\phi(x)\|_2^2\notag\\
     & \leq \kappa^2\sum_{j=1}^{c}\big|\psi^{(j)}(h^{\bw}(x),y)\big|^2 \leq \kappa^2\Big(\sum_{j=1}^{c}\big|\psi^{(j)}(h^{\bw}(x),y)\big|\Big)^2
     \leq L^2\kappa^2.\label{sgd-1}
  \end{align}
  According to the SGD update, we know
  \begin{align*}
    \big\|\bw_{t+1}\big\|_{2,2} & =\Big\|\bw_t-\eta_t\big(\ell'(\bw_t;z_{i_t})+\sigma\bw_t\big)\Big\|_{2,2} \\
     & \leq \big(1-\eta_t\sigma\big)\|\bw_t\|_{2,2}+\eta_t\|\ell'(\bw_t;z_{i_t})\|_{2,2}\\
     & \leq \big(1-\eta_t\sigma\big)\|\bw_t\|_{2,2}+\eta_t\sigma(L\kappa/\sigma)\\
     & \leq \max\{\|\bw_t\|_{2,2},L\kappa/\sigma\},
  \end{align*}
  where we have used \eqref{sgd-1} and $\eta_t\leq1/\sigma$. Applying this inequality recursively, we know by induction that
  \[
  \|\bw_{t+1}\|_{2,2}\leq \Big\{\|\bw_1\|_{2,2},\ldots,\|\bw_t\|_{2,2,},L\kappa/\sigma\Big\}=L\kappa/\sigma.
  \]
  This together with \eqref{sgd-1} shows that
  \[
  \|f'(\bw_t;z)\|_{2,2}\leq\|\ell'(\bw_t;z)\|_{2,2}+\sigma\|\bw_t\|_{2,2}\leq 2L\kappa.
  \]
  Note $r(\bw)$ is $\sigma$-strongly convex w.r.t. $\|\cdot\|_{2,2}$. Therefore, assumptions in Lemma \ref{lem:sgd} hold. We can apply Lemma \ref{lem:sgd} to show the following inequality with probability at least $1-\delta/2$
  \begin{equation}\label{sgd-2}
    F_S(\bw_T)-\inf_{\bw}F_S(\bw)=O\Big(\frac{\log(T)\log(1/\delta)}{T\sigma}\Big).
  \end{equation}
  According to \eqref{main-a}, with probability at least $1-\delta$ there holds
  \begin{align*}
    F(\bw_T)-F(\bw^*) &=  \max\{F_S(\bw_T)-F_S(\bw^*),1/(n\sigma)\}+ \widetilde{O}\Big(\Big(\frac{\big(F(\bw_T)-F(\bw^*)\big)\big(\log(1/\delta)+\widetilde{B}\big)}{n\sigma}\Big)^{\frac{1}{2}}\Big)\\
    & = O\Big(\frac{\log(T)\log(1/\delta)}{\min\{T,n\}\sigma}\Big)+\widetilde{O}\Big(\Big(\frac{\big(F(\bw_T)-F(\bw^*)\big)\big(\log(1/\delta)+\widetilde{B}\big)}{n\sigma}\Big)^{\frac{1}{2}}\Big)\\
  \end{align*}
  where we have used $F_S(\bw^*)\geq\inf_{\bw}F_S(\bw)$ and \eqref{sgd-2} (It can be shown $\widetilde{B}\leq\kappa$). Solving the above quadratic inequality of $F(\bw_T)-F(\bw^*)$ yields the stated bound \eqref{thm-sgd-res} with probability at least $1-\delta$. The proof is complete.
\end{proof}

\subsection{Proof of Proposition \ref{prop:loss-label}\label{sec:loss-label}}
\begin{proof}[Proof of Proposition \ref{prop:loss-label}]
  We first prove the first part.
  For any $\bw,\bw'\in\wcal$, by the elementary inequality
  \[
  \big|\max\{a_1,\ldots,a_c\}-\max\{b_1,\ldots,b_c\}\big|\leq \max\{|a_1-b_1|,\ldots,|a_c-b_c|\}
  \]
  we have
  \begin{align*}
     \big|\ell(\bw;z)-\ell(\bw';z)\big|
     & = \Big|\max_{j\in[c]}\tilde{\ell}\big(y^{(j)}\langle\bw_j,\phi(x)\rangle\big)-\max_{j\in[c]}\tilde{\ell}\big(y^{(j)}\langle\bw_j',\phi(x)\rangle\big)\Big| \\
     & \leq \max_{j\in[c]}\big|\tilde{\ell}\big(y^{(j)}\langle\bw_j,\phi(x)\rangle\big)-\tilde{\ell}\big(y^{(j)}\langle\bw_j',\phi(x)\rangle\big)\big| \\
     & \leq L\max_{j\in[c]}\big|\langle\bw-\bw_j',\phi(x)\rangle\big|=L\big\|h^{\bw}(x)-h^{\bw'}(x)\big\|_\infty.
  \end{align*}
  This proves the first part.

  We now turn to the second part. For any $\bw,\bw'\in\wcal$
  \begin{align*}
    \big|\ell(\bw;z)-\ell(\bw';z)\big|
    &= \frac{1}{|y_+||y_-|}\Big|\sum_{j_+\in y_+}\sum_{j_-\in y_-}
    \Big(\tilde{\ell}\big(\langle\bw_{j_+}-\bw_{j_-},\phi(x)\rangle\big)
    -\tilde{\ell}\big(\langle\bw'_{j_+}-\bw'_{j_-},\phi(x)\rangle\big)\Big)\Big|\\
    & \leq \max_{j_+\in y_+,j_-\in y_-}\Big|\tilde{\ell}\big(\langle\bw_{j_+}-\bw_{j_-},\phi(x)\rangle\big)-\tilde{\ell}\big(\bw'_{j_+}-\bw'_{j_-},\phi(x)\rangle\big)\Big|\\
    & \leq \frac{L}{2}\max_{j_+\in y_+,j_-\in y_-}\Big|\langle\bw_{j_+}-\bw_{j_-},\phi(x)\rangle-\langle\bw'_{j_+}-\bw'_{j_-},\phi(x)\rangle\Big|\\
    & \leq \frac{L}{2}\Big(\max_{j_+\in y_+}\big|\langle\bw_{j_+}-\bw'_{j_+},\phi(x)\rangle\big|+\max_{j_-\in y_-}\big|\langle\bw_{j_-}-\bw'_{j_-},\phi(x)\rangle\big|\Big)\\
    & \leq L\|h^{\bw}(x)-h^{\bw'}(x)\|_\infty.
  \end{align*}
  This proves the second part and finishes the proof.
\end{proof}

\newpage
\section{Some Tables}
\begin{table*}[htbp]
\caption{Generalization bounds for MLC in a Frobenius learning framework (either constraint $\|W\|_F\leq 1$ or regularizer $\|W\|_F^2$)\label{tab:mlc}}
\centering
\begin{tabular}{|c|c|c|c|c|}
  \hline
  Bound & Lipschitz Continuity & Additional Assumption & Method & Reference \\ \hline
  $\sqrt{c/n}$ & $\ell_1$-norm & sub-gaussian distribution  & ERM & \citet{yu2014large} \\ \hline
  $\sqrt{c/n}$ & $\ell_2$-norm & No  & ERM & \citet{liu2018dual,wu2020multi} \\ \hline
  $\sqrt{c}/n$ & $\ell_2$-norm & decay of singular values  & ERM &  \citet{li2019learning} \\ \hline
  $\log^3(nc)/n$ & $\ell_\infty$-norm & strong convexity & RRM/SGD &  This work \\
  \hline
\end{tabular}
\end{table*}
\begin{table*}[htbp]
\caption{Description of the datasets. $c$ is the number of classes, $n$ is the sample size and $d$ is the input dimension.
		\label{tab:data_set}}
\small
\centering
  \begin{tabular}{*{4}{|c}|}\hline
  Dataset & $c$ & $n$ & $d$ \\\hline
  CIFAR10 & $10$ & $10,000$ & $3,072$ \\\hline
  RCV1 & $53$ & $15,564$  & $47,236$ \\\hline
  SECTOR & $105$ & $3,207$ & $55,197$ \\\hline
  ALOI & $1,000$ & $108,000$ & $128$ \\\hline
  \end{tabular}    		
\end{table*}
\end{document}